%% file: main.tex
\PassOptionsToPackage{table,xcdraw,x11names}{xcolor}
\documentclass[10pt,twocolumn,letterpaper]{article}

\usepackage[pagenumbers]{cvpr} %

\input{preamble}

\definecolor{cvprblue}{rgb}{0.21,0.49,0.74}
\usepackage[pagebackref,breaklinks,colorlinks,citecolor=cvprblue]{hyperref}
\usepackage{multirow}
\usepackage[normalem]{ulem}
\usepackage{verbatim}
\usepackage{comment}
\usepackage[accsupp]{axessibility}

\newcommand{\RN}[1]{
  \hspace{-0.2em}{\textbf{\romannumeral#1})}\hspace{-0.4em}
}

\newcommand{\greencell}{\cellcolor{Green2!25}}
\newcommand{\orangecell}{\cellcolor{Orange1!40}}
\newcommand{\redcell}{\cellcolor{IndianRed1!40}}
\useunder{\uline}{\ul}{}
\definecolor{yellow}{rgb}{1, 1, 0.7}
\definecolor{orange}{rgb}{1, 0.85, 0.7}
\newcommand\customparagraph[1]{\vspace{0.4em}\noindent\textbf{#1.}}

\title{GLACE: Global Local Accelerated Coordinate Encoding}

\author{Fangjinhua Wang$^1$\footnotemark[1]
        \quad
        Xudong Jiang$^1$\footnotemark[1]
        \quad
        Silvano Galliani$^2$
        \quad
        Christoph Vogel$^2$
        \quad
        Marc Pollefeys$^{1,2}$\\
        $^1$Department of Computer Science, ETH Zurich\\
        $^2$Microsoft Mixed Reality \& AI Zurich Lab}

\begin{document}
\maketitle

{
    \renewcommand{\thefootnote}%
        {\fnsymbol{footnote}}
        \footnotetext[1]{Equal contribution.}
}

\input{sec/0_abstract}    
\input{sec/1_intro}

\input{sec/2_relatedwork}
\input{sec/3_method}

\input{sec/4_experiment}
\input{sec/5_conclusion}
\input{sec/X_suppl}

\clearpage
{
    \small
    \bibliographystyle{ieeenat_fullname}
    \bibliography{main}
}

\end{document}

%% file: preamble.tex
\usepackage[dvipsnames]{xcolor}

%% file: sec/0_abstract.tex
\begin{abstract}
Scene coordinate regression (SCR) methods are a family of visual localization methods that directly regress 2D-3D matches for camera pose estimation. 
They are effective in small-scale scenes but face significant challenges in large-scale scenes that are further amplified in the absence of ground truth 3D point clouds for supervision. 
Here, the model can only rely on reprojection constraints and needs to implicitly triangulate the points.
The challenges stem from a fundamental dilemma:
The network has to be invariant to observations of 
the same landmark at different viewpoints and lighting conditions, \etc, 
but at the same time discriminate unrelated but similar observations. 
The latter becomes more relevant and severe in larger scenes. 
In this work, we tackle this problem by introducing the concept of co-visibility to the network.
We propose GLACE, which integrates pre-trained global and local encodings and enables SCR to scale to large scenes with only a single small-sized network.
Specifically, we propose a novel feature diffusion technique that implicitly groups the reprojection constraints with co-visibility and avoids overfitting to trivial solutions. Additionally, our position decoder parameterizes the output positions for large-scale scenes more effectively. Without using 3D models or depth maps for supervision, our method achieves state-of-the-art results on large-scale scenes with a low-map-size model. On Cambridge landmarks, with a single model, we achieve 17\% lower median position error than Poker, the ensemble variant of the state-of-the-art SCR method ACE. Code is available at: \url{https://github.com/cvg/glace}.

\end{abstract}

%% file: sec/1_intro.tex
\section{Introduction}
\label{sec:intro}

Visual localization describes the task of estimating the camera position and orientation for a query image in a known scene. This ability to localize in the environment is fundamental and important for applications like robotics, autonomous driving, and Augmented / Virtual Reality. 

\begin{figure}[t]
  \centering
  \setlength{\belowcaptionskip}{-0.5cm}
   \includegraphics[width=1.0\linewidth]{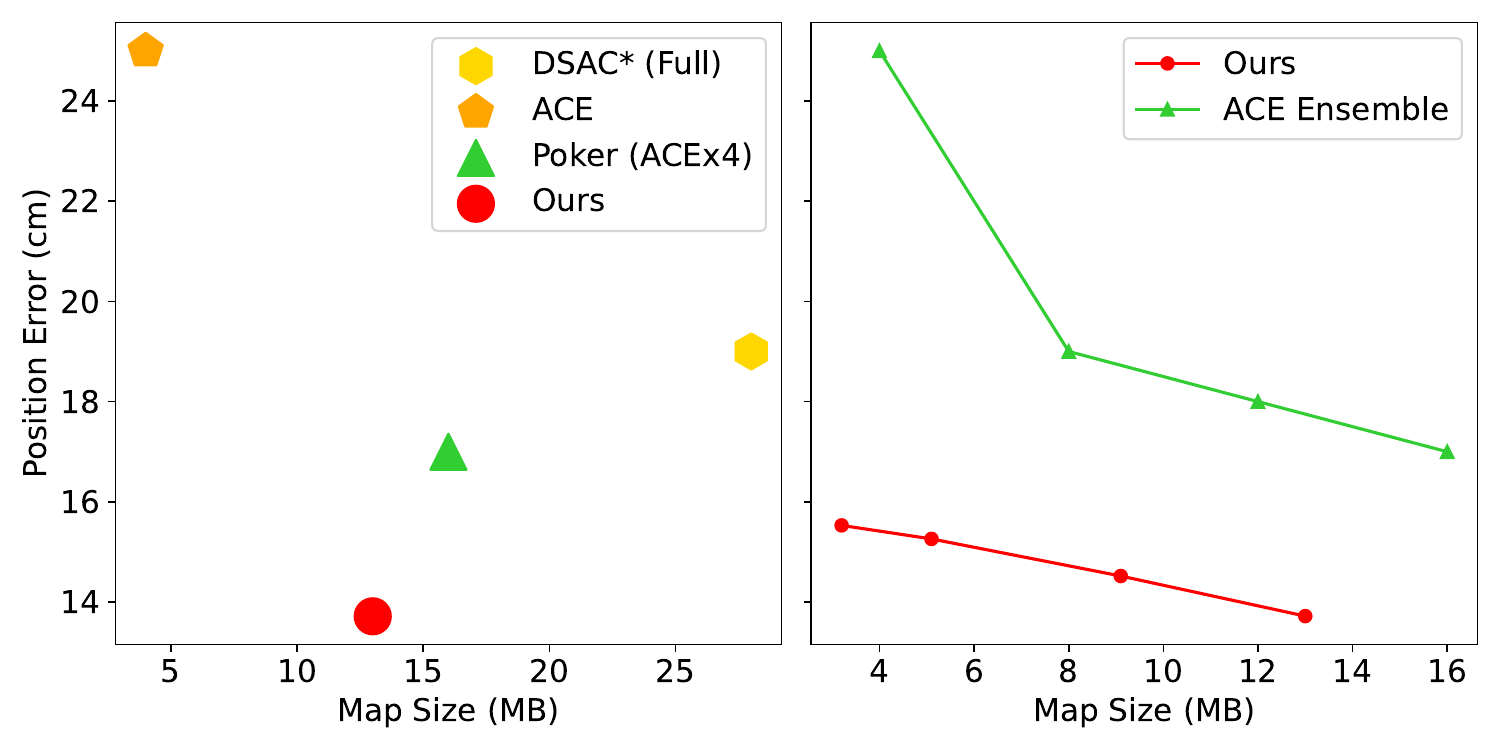}
   \caption{Left: Quantitative comparison of map size and position error with state-of-the-art SCR methods~\cite{brachmann2017dsac,brachmann2023accelerated} on Cambridge landmarks~\cite{kendall2015posenet}. Our method outperforms DSAC~\cite{brachmann2017dsac}, ACE~\cite{brachmann2023accelerated} and Poker (4 ACE models) with a moderate model size. Right: Relationship between map size and position error. Note that our method with the smallest map size (3.2 MB) still performs better than Poker (4 ACE models, map size is 16.0 MB). }
   \label{fig:teaser}
\end{figure}

Currently, most state-of-the-art localization methods are structure-based~\cite{brachmann2017dsac, Brachmann2019ESAC,sarlin2019coarse,sarlin2020superglue,sattler2016efficient,taira2018inloc,toft2020long}, including feature matching based methods and most scene coordinate regression methods. 
Both techniques have in common to build maps from images with known poses. 
For localization they establish matches between 2D pixel positions in the query image and 3D points in the maps. Finally, embedded into RANSAC~\cite{barath2019magsac,barath2019magsacplusplus}, a Perspective-n-Point (PnP) solver~\cite{haralick1994review,bujnak2008general} is used to predict the camera pose from the 2D-3D correspondences. 
Both methodologies differ in the representation of the map and the estimation of correspondences.

Given a database of images, methods based on feature matching~\cite{sarlin2019coarse,sarlin2020superglue,sattler2016efficient,taira2018inloc,taira2019right,schonberger2018semantic} typically represent the 3D scene by reconstructing the 3D geometry,~\eg point cloud, using structure-from-motion (SfM)~\cite{schoenberger2016sfm}.  At test time, they establish 2D-3D matches between pixels in a query image and 3D points in the 3D model using descriptor matching. However, these methods need to store point-wise visual descriptors for the whole point cloud, which may cause storage issues when the scenes scale up. 

In contrast, scene coordinate regression (SCR) methods~\cite{brachmann2023accelerated,brachmann2017dsac,Cavallari2019network,Brachmann2018dsacpp,dong2022visual} implicitly encode the map information inside a deep neural network. Instead of computing 2D-3D matches via explicit descriptor matching, these methods directly regress the matches. 
Though achieving superior performance in small scenes~\cite{shotton2013scene}, it is difficult to scale these methods to large-scale scenes due to the limited capacity of a single network~\cite{brachmann2017dsac}. A common solution is to train multiple networks on sub-regions of the scene~\cite{Brachmann2019ESAC}. But this certainly increases the model size, training time, and query time. 
Recent works~\cite{brachmann2023accelerated,Brachmann2018dsacpp,Brachmann2021dsacstar} avoid the need for depth maps or a complete 3D model for training. In addition, ACE~\cite{brachmann2023accelerated} proposes a method to train a 4MB-sized network in 5 minutes, while achieving state-of-the-art performance for smaller scenes~\cite{shotton2013scene}. Although it has impressive efficiency, ACE~\cite{brachmann2023accelerated} still possesses the same problem of scaling to larger problem sizes and requires the use of an ensemble of networks for large-scale scenes~\cite{kendall2015posenet}, which lessens efficiency and practicality. 

In this work, we propose GLACE, a novel method that enables the scene regression methodology to work on large-scale scenes with only a single network. Our method achieves state-of-the-art results on several large-scale datasets~\cite{kendall2015posenet,Brachmann2019ESAC} while using only a single model of small size and without using 3D models for supervision. 
Our contribution can be summarized as follows: 

\RN{1} To our knowledge, our method is the first attempt of an SCR method to achieve state-of-the-art performance on large-scale scenes without using an ensemble of networks or 3D model supervision. 

\RN{2} We propose a novel feature diffusion technique for the pre-trained global encodings that implicitly groups the reprojection constraints with co-visibility, which avoids overfitting to trivial solutions. 

\RN{3} We propose a positional decoder that parameterizes the output positions for large-scale scenes more effectively than previous work.

%% file: sec/2_relatedwork.tex
\section{Related Work}
\label{sec:related_work}

\customparagraph{Pose Regression}
 Pose regression approaches~\cite{kendall2015posenet,Kendall2017GeometricLF,Brahmbhatt2018mapnet,Shavit2021MStransformer,turkoglu2021visual,WinkelbauerICRA21,zhou2020essnet,naseer2017deep} encode the scene into a neural network and are trained end-to-end. At test time they regress an absolute or relative pose from a query image. 
Without geometric constraints, absolute pose regression methods~\cite{kendall2015posenet,Kendall2017GeometricLF,naseer2017deep,Walch2017lstm} usually do not generalize well to novel viewpoints or appearances. Besides, these methods do not scale well when limiting network capacity~\cite{taira2018inloc}. 
Operating differently, relative pose regression methods~\cite{balntas2018relocnet,ding2019camnet,zhou2020essnet} regress a camera pose relative to one or more database images. While being scene-agnostic, they are often limited in accuracy. 

\customparagraph{Feature Matching Based Localization}
Localization methods based on feature matching (FM)~\cite{sarlin2019coarse,sarlin2020superglue,sattler2016efficient,taira2018inloc,taira2019right,schonberger2018semantic,svarm2016city,sattler2015hyperpoints} 
are often still considered state-of-the-art for visual localization. 
Those methods establish 2D-3D correspondences between pixels in a query image
and 3D points in a scene model using descriptor matching. To scale to large scenes and handle challenging problems, such as day-night illumination change and seasonal change, these methods first perform a form of coarse localization. For instance, using image retrieval~\cite{netvlad,torii201524}, to first identify a small set of potentially relevant database images and only then perform descriptor matching with the 3D points visible in these images. 
However, these methods need to store all the descriptor vectors of the 3D model to perform matching, which may cause storage issues for large maps. Recently, several works~\cite{zhou2022gomatch,Panek2022meshloc} try to avoid storing descriptors explicitly and instead propose to match directly against the geometry,~\eg, given as point cloud or mesh. 

\customparagraph{Scene Coordinate Regression} 
Given a query image, this family of localization methods regresses for a 2D pixel the corresponding 3D coordinates in the scene~\cite{shotton2013scene}. Usually, these methods implicitly store the information about the scene within the weights of a machine learning model. 
To regress 2D-3D matches, SCR methods are mainly based on random forests~\cite{brachmann2016,cavallari2017fly,Cavallari2019cascade,shotton2013scene,valentin2015cvpr} or convolutional neural networks~\cite{brachmann2017dsac,Brachmann2018dsacpp,Brachmann2019ESAC,Brachmann2021dsacstar,brachmann2023accelerated,dong2022visual,li2020hierarchical}.
Recently, ACE~\cite{brachmann2023accelerated} only uses posed RGB images for mapping. The training is performed only from the images using a loss based on the image reprojection error while completely avoiding the explicit reconstruction of a 3D model. It achieves state-of-the-art performance on several small-scale scenes~\cite{shotton2013scene,valentin2016learning}, and demonstrates impressive efficiency in training time and map size. 
However, a single model based on SCR is usually limited to only working on scenes of small-scale~\cite{brachmann2017dsac}. 
Larger scenes require techniques like an ensemble of SCR networks~\cite{Brachmann2019ESAC,brachmann2023accelerated} to scale, 
which demands additional maintenance, training time, and memory. 
In contrast, our method scales SCR methods to large-scale scenes without requiring an ensemble of networks or 3D model supervision. 

%% file: sec/3_method.tex
\section{Method}\label{sec:method}
In this section, we first introduce the basic concepts for scene coordinate regression with ACE~\cite{brachmann2023accelerated}.
We follow by discussing how the system performs implicit triangulation when training without ground truth scene coordinate supervision
and describe challenges in large-scale scenes for SCR methods.
We conclude by introducing co-visibility to SCR in the form of global encodings 
and explain how we effectively enable the network to utilize this information. 
Finally, we discuss our novel position decoding technique that removes a bias in the SCR 
toward producing solutions near the center of training camera positions.
\begin{figure}
    \centering
    \includegraphics[width=\linewidth]{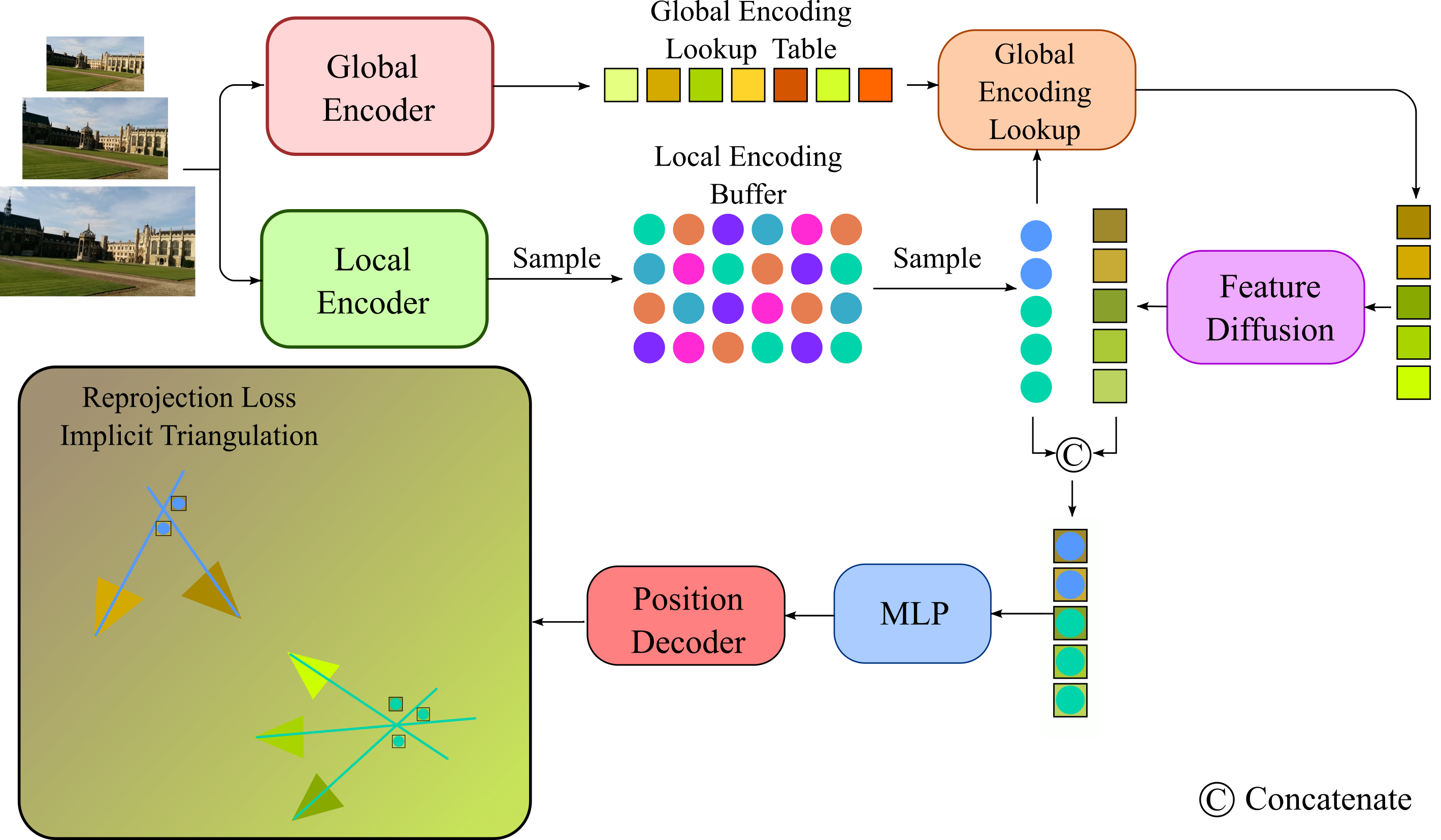}
    \caption{Pipleine of GLACE. Besides the buffer of ACE~\cite{brachmann2023accelerated} local encodings, we extract global features of training images with image retrieval model~\cite{r2former}. During training, we sample a batch of local encodings, look up their global encoding according to their image index and perform feature diffusion by adding Gaussian noise. The global and local encodings are concatenated as input to an MLP head. The output of the MLP is further processed by a position decoder to yield the final coordinate predictions. The global encoding with feature diffusion facilitates the grouping of reprojection constraints, enabling effective implicit triangulation in large-scale scenes. Best viewed when zoomed in. }
    \label{fig:pipeline}
\end{figure}

\subsection{Scene Coordinate Regression}
\label{sec:SCR}
\customparagraph{Visual Localization}
We consider visual localization from RGB images. 
For training, we require a set of images with corresponding ground truth poses$\{(I_{train}, h_{train} )\}$, where $h_{train}$ denotes the rigid transformation from world coordinates to camera coordinates. During testing, our system estimates the camera pose $h_{test}$ for a query image $I_{test}$. To that end, we follow the SCR methodology. Specifically, we mainly consider the setting with a pretrained local feature extractor and without ground truth scene coordinate supervision, established by ACE~\cite{brachmann2023accelerated}. We first briefly review the SCR pipeline.

\customparagraph{SCR Pipeline}
 SCR methods belong to the structure-based methods, which first predict 2D-3D correspondences and then solve for the pose with PnP and RANSAC. Traditional structure-based methods usually explicitly store a triangulated point cloud with corresponding features and match them with query image features to obtain 2D-3D correspondences. Instead, SCR methods implicitly learn the 2D-3D correspondence, usually in a convolutional neural network, which outputs the corresponding 3D coordinate for each image patch:
\begin{equation}
    y_i = f(p_i),
\end{equation}
where $p_i$ is the image patch centered at pixel $x_i$ and $y_i$ is the corresponding 3D coordinate, the function $f$ is given by the neural network.
Previous works~\cite{brachmann2017dsac,Brachmann2019ESAC,hscnetpp,li2020hierarchical} supervise the output $y_i$ by providing ground truth 3D scene coordinates, \eg, from a depth sensor or an SfM point cloud. 

\customparagraph{SCR with Reprojection Loss}
Some recent works~\cite{Brachmann2018dsacpp,brachmann2023accelerated} enable training without ground truth scene coordinates by employing a reprojection loss:
\begin{equation}
    e_{\pi}(x_i,y_i,h) = || x_i- \pi(\mathbf{K} \cdot h \cdot y_i)||_1,
\end{equation}
where $h$ is the ground truth pose of the image $\mathbf{K}$ is the camera intrinsic matrix and $\pi$ performs the mapping from homogeneous to pixel space.
The reprojection loss is usually combined with a robust loss function to reduce the influence of outliers. We use the dynamic tanh loss introduced in ACE~\cite{brachmann2023accelerated}:
\begin{equation}\label{eq:robust_loss}
    l_{\pi}(x_i,y_i,h)=\begin{cases}
        \tau (t) \tanh(\frac{e_{\pi}(x_i,y_i,h)}{\tau (t)}), & \text{if } y_i \in V \\
        ||y_i - \bar{y}_i||_1, & \text{otherwise}
        \end{cases}   
\end{equation}
where $V$ is the set of valid predictions, defined as points that are between 0.1m to 1000m in front of the camera and have a reprojection error $e_{\pi}(x_i,y_i,h)$ less than 1000px. $\bar{y}_i$ is the pseudo ground truth scene coordinate defined by the inverse projection of the pixel with the ground truth pose and a fixed target depth at 10m. 
During training the threshold
$\tau (t)$ is adjusted dynamically based on the relative training time $t$:
\begin{equation}
    \tau (t) =  \sqrt{1-t^2} \tau_{max} + \tau_{min}.
\end{equation}

\customparagraph{Reprojection Loss as Implicit Triangulation}
In standard reconstruction, 2D-3D correspondences are explicitly established through matching. Observations of the same 3D point are grouped into a track, and the 3D point is triangulated by minimizing their reprojection error. 
In contrast, in SCR methods such as ACE~\cite{brachmann2023accelerated} and ours, there is no explicit grouping of 2D observations for the same 3D point. 
Instead, each 2D observation \textit{independently} regresses to a 3D point. Though initially seems like an under-determined problem, these methods demonstrate practical efficacy, which we attribute to an \emph{implicit triangulation} process. This process is driven by the inherent prior of neural networks to deliver smooth functions~\cite {pmlr,tancik2020fourfeat}, where similar inputs tend to produce similar outputs and undergo similar supervision. Thus, the reprojection loss for similar inputs is collectively minimized, leading to the triangulation of their corresponding output points. This insight explains the practical success of such methods, but also underlines the problem of applying SCR on large maps, which possess unrelated, yet, visually similar image observations and provides the motivation for our feature diffusion techniques.

\subsection{Global Local Encoding}
\customparagraph{Challenges in Large-scale Scenes}
SCR methods possess state-of-the-art accuracy in small indoor scenes. However, they struggle in larger environments, especially when no ground truth scene coordinate supervision is available, and the network needs to perform implicit triangulation of the coordinates from scratch. Consider the trade-off between invariance and discriminative power. Specifically, the dilemma of the receptive field. A model with a smaller receptive field satisfies the invariance assumption better and can generalize to new observations, but suffers from ambiguity when there are different locations with similar local appearances, which intuitively occurs more frequently in large-scale scenes. On the other hand, a model with a larger receptive field may be able to disambiguate similar patches in different locations, but this breaks the invariance assumption: the network will also distinguish observations of the same scene coordinate. This can lead to overfitting to trivial solutions, \eg, producing an arbitrary point along the ray, instead of triangulating the point from different observations. 
This also leads to poor generalization, since novel views of a point observed in training cannot be associated with it anymore. 

\begin{figure}
    \centering
    \includegraphics[width=\linewidth]{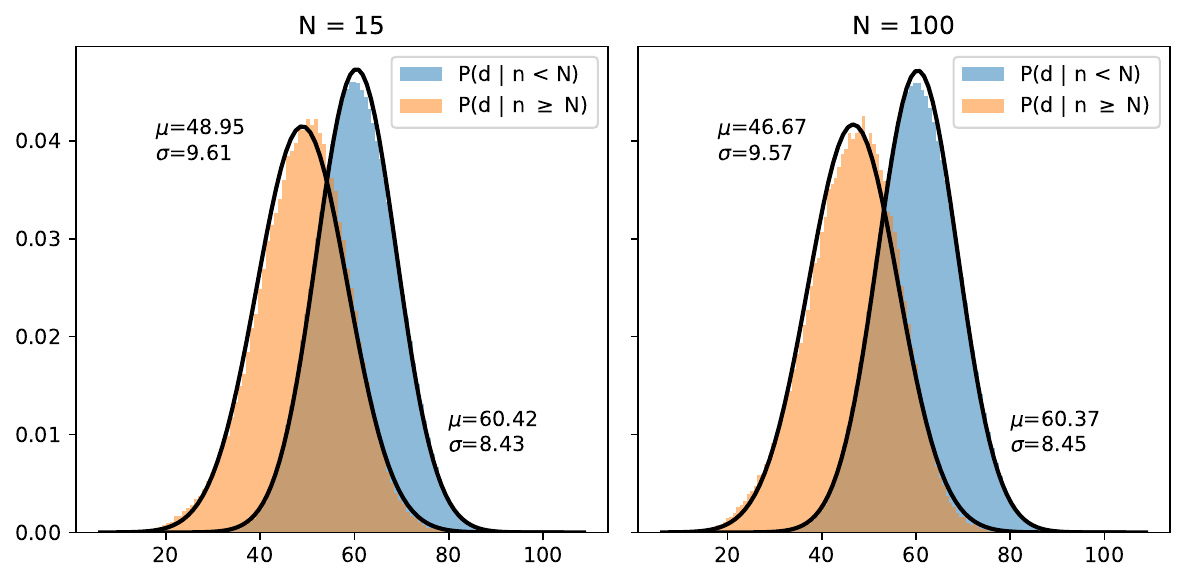}
    \caption{Distribution of angular feature distance($^\circ$), conditioned on co-visibility. Two images are considered co-visible, if the number of co-visible points $n$ at least reaches a threshold $N$. The x-axis depicts the angular  distance $d$ in degrees (\emph{left}: N=15, \emph{right}: N=100). }
    \label{fig:dist|covis}
\end{figure}

\begin{figure}
    \centering
    \includegraphics[width=\linewidth]{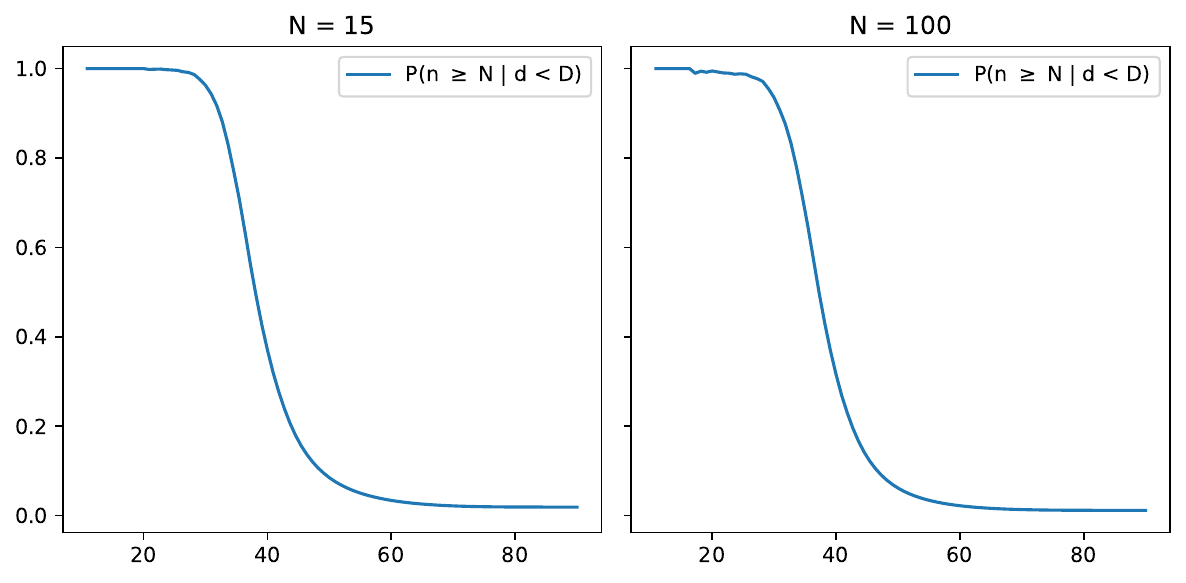}
    \caption{Distribution of co-visibility conditioned on the angular feature distance($^\circ$). Two images are considered co-visible, if the number of co-visible points $n$ at least reaches a threshold $N$. The x-axis depicts the angular threshold $D$ (\emph{left}: N=15, \emph{right}: N=100). }
    \label{fig:covis|dist}
\end{figure}

\customparagraph{Global Encoding}
In order to solve the dilemma, we propose to carefully introduce global information, only including what is necessary. First, we analyze what exactly is needed from global information. 
Without global information, 
ambiguous patches that may belong to different scene points will together, via the reprojection loss, affect the triangulation of the same point. 
Also, the robustness in the loss function Eq.~\ref{eq:robust_loss} can only mitigate, but not solve such problems. 
Therefore, we need global information to effectively group the reprojection constraints. Specifically, we only want to triangulate points in two images if and only if they are looking at the same thing,~\ie the views share sufficient co-visible structure.
To effectively measure co-visibility, we utilize a global feature from an image retrieval model $R^{2}$Former~\cite{r2former}, pretrained on MSLS dataset~\cite{MSLSdataset} and supervised by triplet margin loss with margin $m$:
\begin{equation}
    L_{retrieval} = \text{max}(|| E_q - E_p||^2 - ||E_q- E_n||^2 +m ,0),
    \label{eq:retrievalloss}
\end{equation}
where $E_q, E_p, E_n$ are global features of query, positive, and negative samples. The global features are 256-dimensional vectors normalized to the unit sphere. Here, we further analyze the relationship between feature distance and co-visibility using the SfM reconstruction of a large scene. 
A generative modeling analysis in~\cref{fig:dist|covis} depicts the distribution of the angular feature distance($^\circ$) $d =\frac{180}{\pi} \arccos(u\cdot v)$ conditioned on the number of co-visible points. 
It strongly reminds us of a mixture of Gaussians, where the distribution of co-visible pairs possesses a lower mean. 
The discriminative model in Fig.~\ref{fig:covis|dist} shows that the conditional probability of co-visibility $c$ conditioned on feature distance $d$ resembles $ P(c|d) \sim \text{Bernoulli}(p)$, where the parameter $p$ equals a sigmoid-like function of the feature distance. $p$ is high before a 'threshold', then starts to decrease quickly to a low level afterward, which implies that it is possible to discriminate co-visibility based on the feature distance.

\customparagraph{Naive Concatenation}
With the co-visibility information contained in the global encoding, we still need to effectively integrate global and local features. 
First, consider the naive concatenation. 
In our discussion above, we assume that patches with similar input encoding will triangulate the same point together. 
When we concatenate local and global encoding together, inputs will triangulate the same point when both the local and global encoding are similar. 
However, as shown in Fig.~\ref{fig:dist|covis}, the feature distance between co-visible image pairs, although generally smaller than non-covisible pairs, may still be quite large. 
Intuitively, views of the same point with some angle between them will only partially overlap and thus possess global descriptors that do not match as well as local descriptors, and the concatenated descriptors of matching patches will not have a small distance as before. 
Those images that have almost the same global encoding, possess a very similar pose, with a small baseline, and contribute only little to the triangulation. 
Hence, if we simply concatenate global local encodings together, only a few images with small baselines are grouped together, which leads to large triangulation errors. 
Furthermore, the network might struggle to associate unseen views (\wrt to spatial coverage) during testing and generalize badly. 

\customparagraph{Explicit Clustering}
A simple idea to solve this problem is to explicitly cluster the global features, associate each feature with its cluster center, and use this as global encoding. This forces a grouping into 'hard' clusters of features with the same global encoding.
However, this hard clustering approach requires to decide on an appropriate number of clusters. 
The number has to be large enough 
to ensure each cluster has a sufficient number of observations per point for triangulation
and small enough to 
avoid ambiguous local encodings within a cluster, as shown in Tab.~\ref{tab:gfeat-ablation}. 

\customparagraph{Implicit Grouping with Feature Diffusion}
We propose a novel feature diffusion technique to perform the grouping implicitly.
The idea is simple: instead of using a single fixed global feature for each image, 
we add some noise to make it a distribution. 
For the simplicity of sampling, we add Gaussian noise with a standard deviation of $\sigma = m$, where $m$ is the margin for the image retrieval loss in Eq.~\ref{eq:retrievalloss}.
After adding the noise, the encoding is mapped back to the unit sphere.
This method can be viewed as a form of feature metric data augmentation that imposes a stronger smoothness prior on global encoding, which prevents the neural network from easily discriminating co-visible pairs, thereby promoting implicit triangulation. Distinct from traditional image metric augmentations that typically involve alterations in the input image space, such as color jittering. Our approach operates directly within the feature space, where distances more accurately reflect covisibility relationships. The choice of hyperparameters, grounded in the metric space properties of the pretrained encoder, eliminates the need for scene-specific tuning, thereby ensuring robust performance across different scenes.

\begin{table*}[h]
  \centering
  \footnotesize
  \begin{tabular}{clccccccc}
  \toprule
  \multicolumn{1}{l}{} &  & & & \multicolumn{2}{c}{7 Scenes} & \multicolumn{1}{l}{} & \multicolumn{2}{c}{12 Scenes} \\
  \cmidrule{5-6} \cmidrule{8-9} 
  \multicolumn{1}{l}{} & & \multirow{-2}{*}{\begin{tabular}[c]{@{}c@{}}{Mapping w/}\\ {Mesh/Depth}\end{tabular}}  & \multirow{-2}{*}{\begin{tabular}[c]{@{}c@{}}{Map}\\ {Size}\end{tabular}} & {SfM poses} & {D-SLAM poses} & & {SfM poses} & {D-SLAM poses} \\
  \midrule
   & AS (SIFT) \cite{sattler2016efficient} & \greencell No   & \redcell $\sim$200MB & 98.5\% & 68.7\% & & 99.8\% & 99.6\% \\
   & D.VLAD+R2D2 \cite{kapture2020}  & \greencell No   & \redcell $\sim$1GB & 95.7\% & 77.6\% & & 99.9\% & 99.7\% \\
   & hLoc (SP+SG) \cite{sarlin2019coarse, sarlin2020superglue} & \greencell No   & \redcell $\sim$2GB & 95.7\% & 76.8\% & & 100\% & 99.8\% \\
  \multirow{-4}{*}{\rotatebox{90}{{FM}}} 
   & pixLoc \cite{sarlin21pixloc} & \greencell No  & \redcell $\sim$1GB & N/A & 75.7\% & & N/A & N/A \\
  \midrule
   & DSAC* (Full) \cite{Brachmann2021dsacstar} & \redcell Yes  & \greencell 28MB & 98.2\% & 84.0\% & & 99.8\% & 99.2\% \\
   & DSAC* (Tiny) \cite{Brachmann2021dsacstar} & \redcell Yes  & \greencell 4MB & 85.6\% & 70.0\% & & 84.4\% & 83.1\% \\
   & SANet \cite{yang2019sanet} & \redcell Yes  & \redcell $\sim$550MB & N/A & 68.2\% & & N/A & N/A \\
  \multirow{-4}{*}{\rotatebox{90}{\begin{tabular}[c]{@{}c@{}}{SCR} \\ {(w/ Depth)}\end{tabular}}} 
  & SRC \cite{dong2022visual} & \redcell Yes  & \orangecell 40MB & 81.1\% & 55.2\% & & N/A & N/A \\
  \midrule\midrule
  & DSAC* (Full) \cite{Brachmann2021dsacstar} & \greencell No  & \orangecell 28MB & {\ul 96.0\%} & {\ul 81.1\%} & & 99.6\% & {\ul 98.8\%} \\
  & DSAC* (Tiny) \cite{Brachmann2021dsacstar} & \greencell No  & \greencell 4MB & 84.3\% & 69.1\% & & 81.9\% & 81.6\% \\
  \multirow{-2}{*}{\rotatebox{90}{{SCR}}} 
  & ACE~\cite{brachmann2023accelerated} & \greencell No &  \greencell 4MB & \textbf{97.1\%} & 80.8\% & \textbf{} & {\ul 99.9\%} & \textbf{99.6\%} \\
  & GLACE (Ours) & \greencell No &  \greencell 9MB &  95.6\% & \textbf{81.4\%} & & \textbf{100\%} & \textbf{99.6\%} \\
  \bottomrule
  \end{tabular}
  \caption{\textbf{Quantitative results for single scene relocalization.} We report the percentage of frames below a 5cm,5$^\circ$ pose error. For the ``SCR'' group, best results in \textbf{bold}, second best results \underline{underlined}. We list the map size and whether depth (rendered or measured) is needed for mapping.}
  \label{tab:results_single}
  \vspace{-1em}
  \end{table*}

\subsection{Position Decoding}
\begin{figure}[t]
  \centering
   \includegraphics[width=1.0\linewidth]{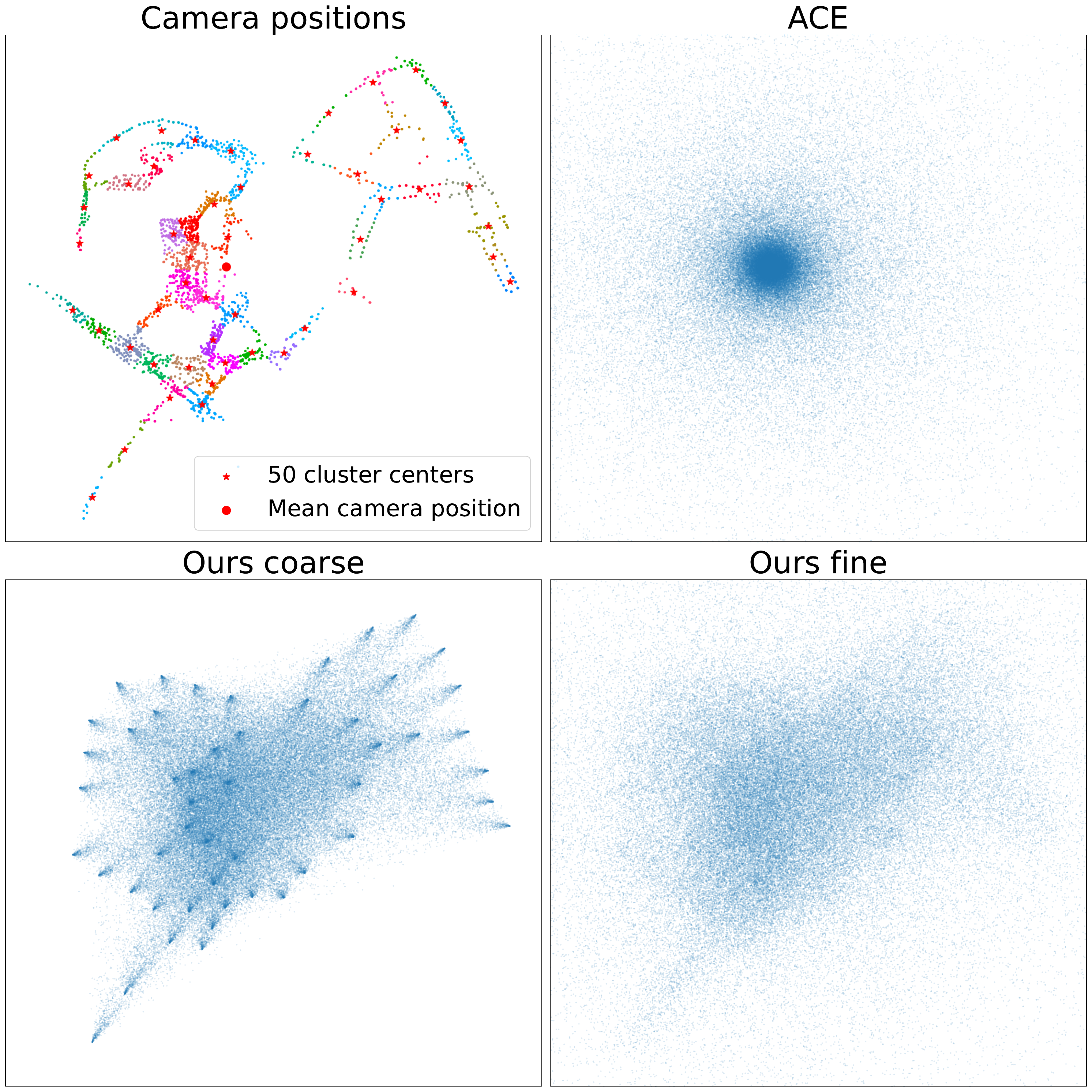}
   \caption{Comparison between decoder output of random Gaussian input samples. We use 50 cluster centers in this example of the Aachen dataset, shown in the top left (cluster assignments are color-coded, and cluster centers occur as \emph{red star}).}
   \label{fig:decodersample}
\end{figure}
Research~\cite{sattler2019limits} shows that the final layer has an important effect on the prior of CNNs that regress spatial positions, if the direct output of the last linear layer is a linear combination of bases in its weight. 
Therefore, it is important to effectively parameterize the final position by the network output, especially when there is no ground truth scene coordinate supervision, and we rely 
on the prior of the model to perform implicit triangulation.
The network output of ACE~\cite{brachmann2023accelerated} $(\Dot{d},\hat{w})$ defines an offset in homogeneous coordinates from the center of training camera positions $c$:
\begin{equation}
    \hat{y}=  \frac{\Dot{d}}{w} + c.
    \label{eq:homo}
\end{equation}
\begin{equation}
    w=\min (\frac{1}{S_{\min }}, \beta^{-1}\log ( 1+ \exp(\beta \hat{w})) + \frac{1}{S_{\max} }).
\end{equation}
$S_{\min}, S_{\max}$ are hyperparameters that define minimum and maximum scale and 
$\beta =  \frac{\log{2}}{1- S_{\max}^{-1}}$ is the parameter for the softplus.
It can better parameterize points at different scales, but still suffers from \emph{an unimodal prior}, preferring localization near the center $c$ (\cref{fig:decodersample}, top right). 
Here, we propose an effective position decoder that predicts a convex combination of cluster center positions to replace the fixed center $c$ in Eq.~\ref{eq:homo}. 
We use K-Means to distribute the training camera positions into $k$ clusters with centers $\{c_i\}$. The final linear layer of our MLP outputs $k$ logits $\{s_i\}$, one for each cluster center and one homogeneous coordinate with parameters $\Dot{d},\hat{w}$ to define an offset. 
The final output is calculated similarly to Eq.~\ref{eq:homo}. 
We only replace the center of training camera positions 
with the convex combination (using the softmax of logits) of cluster centers:
\begin{equation}
    \hat{y} = \frac{\Dot{d}}{\hat{w}} + \sum_{i=1}^k \frac{e^{s_i}}{\sum_j e^{s_j}}c_i.
\end{equation}

We demonstrate the idea of our model and compare it to the encoding of \cite{brachmann2023accelerated} in \cref{fig:decodersample}.
We sample from a Gaussian distribution as input and compare the decoded output for different decoders. 
Because of the unimodal prior of the ACE decoder, most of the samples are concentrated at the center.
As a convex combination of clusters centers our model is inherently multimodal, but the samples are still concentrated at the modes. 
After adding the offset, the samples are distributed more evenly (\cref{fig:decodersample}, bottom right). 
Although the output of an MLP may not be a simple Gaussian distribution, this still can show that our decoder can better parameterize the output. 
We also designed a simple toy experiment in supplementary material using a simplified 2D task that predicts the coordinates of the center pixel of a 2D image patch. 
The results show that even with strong supervision, the original decoder cannot regress the coordinate well when the scale is large. In contrast, with the help of our positional decoder, the performance is improved significantly. Please refer to the supplement for details.

%% file: sec/4_experiment.tex
\section{Experiment}\label{sec:experiment}

\begin{table}[] 
  \centering 
  \begin{tabular}{@{}lcccc@{}} 
  \toprule 
  Method & w / Depth  & Size                          & i12& i19\\ \midrule 
  ESAC  \cite{Brachmann2019ESAC} &Yes&336$/$532MB & 97.1\%                               & 88.1\%                       \\  \midrule 
  ACE~\cite{brachmann2023accelerated} &No& 4MB  &        10.3\%	&5.9\%                \\  
  ACE~\cite{brachmann2023accelerated} $\times$ 4&No& 16MB  & 77.4\%	& 36.5\% \\
  ACE~\cite{brachmann2023accelerated} $\times$ 19 &No& 78MB  & 99.3\%  & 90.9\%                   \\  
  GLACE (Ours) &No& 9MB &   99.1\% & 87.0\%     \\  
  \bottomrule 
\end{tabular} 
  \caption{\textbf{Integrated rooms dataset evaluation} with D-SLAM poses. We report the percentage of frames below a 5cm,5$^\circ$ pose error. ESAC uses 12/19 ensembles and has map size 336/532MB respectively.} 
  \label{tab:results_id}
\end{table}

\begin{table}[] 
  \centering 
  \setlength{\belowcaptionskip}{-0.6cm}
  \begin{tabular}{@{}lccc@{}} 
  \toprule 
  Method   & Size & i12 & i19 \\ \midrule 
  ACE~\cite{brachmann2023accelerated} & 4MB & 9.0\%	& 17.0\%  \\  
  ACE~\cite{brachmann2023accelerated} $\times$ 4& 16MB& 76.9\%	& 42.0\% \\
  ACE~\cite{brachmann2023accelerated} $\times$ 19 & 78MB  &99.9\%  & 97.8\%  \\  
  GLACE (Ours) &9MB & 99.5\%   & 93.4\%     \\  
  \bottomrule 
\end{tabular} 
  \caption{\textbf{Integrated rooms dataset evaluation} with SfM poses. We report the percentage of frames below a 5cm, 5$^\circ$ pose error.} 
  \label{tab:results_i}
\end{table}

\begin{table*}[]
  \centering
  \footnotesize
  \begin{tabular}{clcccccccc}
  \toprule
  \multicolumn{1}{l}{} &  & & & \multicolumn{5}{c}{Cambridge Landmarks} \\
  \cmidrule(l){5-9} 
  \multicolumn{1}{l}{} & & \multirow{-2}{*}{\begin{tabular}[c]{@{}c@{}}Mapping w/\\ Mesh/Depth\end{tabular}}  & \multirow{-2}{*}{\begin{tabular}[c]{@{}c@{}}Map\\ Size\end{tabular}} & Court & King's & Hospital & Shop & St. Mary's 
  & \multirow{-2}{*}{\begin{tabular}[c]{@{}c@{}}Average \\ (cm / $^\circ$)\end{tabular}} \\ 
  \midrule
   & AS (SIFT) \cite{sattler2016efficient} & \greencell No   & \redcell $\sim$200MB & 24/0.1 & 13/0.2 & 20/0.4 & 4/0.2 & 8/0.3 & 14/0.2 \\
   & hLoc (SP+SG) \cite{sarlin2019coarse, sarlin2020superglue} & \greencell No   & \redcell $\sim$800MB & 16/0.1 & 12/0.2 & 15/0.3 & 4/0.2 & 7/0.2 & 11/0.2 \\
   & pixLoc \cite{sarlin21pixloc} & \greencell No   & \redcell $\sim$600MB & 30/0.1 & 14/0.2 & 16/0.3 & 5/0.2 & 10/0.3 & 15/0.2\\
   & GoMatch \cite{zhou2022gomatch} & \greencell No   & \orangecell $\sim$12MB & N/A & 25/0.6 & 283/8.1 & 48/4.8 & 335/9.9 & N/A \\
  \multirow{-5}{*}{\rotatebox{90}{FM}} 
   & HybridSC \cite{compression2019cvpr} & \greencell No  & \greencell $\sim$1MB & N/A & 81/0.6 & 75/1.0 & 19/0.5 & 50/0.5 & N/A \\
  \midrule
   & PoseNet17 \cite{Kendall2017GeometricLF} & \greencell No  & \orangecell 50MB & 683/3.5 & 88/1.0 & 320/3.3 & 88/3.8 & 157/3.3 & 267/3.0\\
  \multicolumn{1}{l}{\multirow{-2}{*}{\rotatebox{90}{APR}}}
   & MS-Transformer \cite{Shavit2021MStransformer} & \greencell No  & \orangecell $\sim$18MB & N/A & 83/1.5 & 181/2.4 & 86/3.1 & 162/4.0 & N/A \\
  \midrule
   & DSAC* (Full) \cite{Brachmann2021dsacstar} & \redcell Yes  & \orangecell 28MB & 49/0.3 & 15/0.3 & 21/0.4 & 5/0.3 & 13/0.4 & 21/0.3 \\
   & SANet \cite{yang2019sanet} & \redcell Yes & \redcell $\sim$260MB & 328/2.0 & 32/0.5 & 32/0.5 & 10/0.5 & 16/0.6 & 84/0.8\\
  \multirow{-3}{*}{\rotatebox{90}{\begin{tabular}[c]{@{}c@{}}SCR \\ w/ Depth\end{tabular}}}
  & SRC \cite{dong2022visual} & \redcell Yes  & \orangecell 40MB & 81/0.5 & 39/0.7 & 38/0.5 & 19/1.0 & 31/1.0 & 42/0.7\\
  \midrule
  \midrule
  \multirow{5}{*}{\rotatebox{90}{SCR}}
   & DSAC* (Full) \cite{Brachmann2021dsacstar} & \greencell No  & \orangecell 28MB &  34/0.2 & \textbf{18/0.3} & {\ul 21/0.4} & {\ul 5/0.3} &  {\ul 15/0.6} &  19/0.4 \\
   & DSAC* (Tiny) \cite{Brachmann2021dsacstar} & \greencell No  & \greencell 4MB & 98/0.5 &  27/0.4 & 33/0.6 &  11/0.5 & 56/1.8 & 45/0.8\\
  & ACE~\cite{brachmann2023accelerated}  & \greencell No  & \greencell 4MB & {43/0.2} & 28/0.4 & {31/0.6} & {\ul 5/0.3} & {18/0.6} & 25/0.4\\
  & Poker (ACE~\cite{brachmann2023accelerated} $\times$ 4) & \greencell  No  & \orangecell 16MB & {\ul 28/0.1} & \textbf{18/0.3} &  25/0.5 & {\ul 5/0.3} & \textbf{9/0.3} & {\ul 17/0.3} \\
  & GLACE (ours) & \greencell No  & \orangecell 13MB & \textbf{19/0.1} & {\ul 19/0.3} & \textbf{17/0.4} & \textbf{4/0.2} & \textbf{9/0.3} & \textbf{14/0.3} \\
  \bottomrule
  \end{tabular}
  \caption{\textbf{Cambridge Landmarks \cite{kendall2015posenet} Results.} We report median rotation and position errors. Best results in \textbf{bold} for the ``SCR'' group, second best results \underline{underlined}.}
  \label{tab:results_cam}
  \end{table*}

\subsection{Datasets}

7 Scenes~\cite{shotton2013scene} and 12 Scenes~\cite{valentin2016learning} are two standard datasets for room-scale indoor RGB-D localization. 
They contain 7 and 12 scenes respectively, each with a set of RGB-D sequences. There are two sets of ground truth poses for each scene, one from SfM and one from depth-based SLAM. Since they both have some bias~\cite{brachmann2021limits}, we report results on both of them following prior work~\cite{brachmann2023accelerated}. 
To evaluate localization in large-scale indoor scenes, previous works~\cite{Brachmann2019ESAC,hscnetpp} have proposed to integrate multiple rooms from 7 Scenes and 12 Scenes into a single scene, denoted by i7, i12, and i19. We strictly follow ~\cite{Brachmann2019ESAC}, placing the scenes inside a 2D grid with a cell size of $5 m$. 

Cambridge Landmarks~\cite{kendall2015posenet} is a large-scale outdoor dataset, with RGB sequences of landmarks in Cambridge. It includes ground truth poses and a sparse 3D reconstruction generated via SfM. The dataset is notable for its large-scale and outdoor setting, providing a different set of challenges compared to small-scale indoor datasets. 

The Aachen Day-Night dataset~\cite{sattler2018benchmarking,sattler2012image} is a city-scale dataset, which is particularly challenging for SCR methods due to its large scale and sparsity. It contains only limited images of Aachen city and ground truth poses provided via SfM. Here, we only consider Aachen Day, because there is no night-time training data. 

\subsection{Implementation}
\customparagraph{Architecture}
We implement our method in PyTorch based on the official implementation of ACE~\cite{brachmann2023accelerated}. The MLP architecture is the same as ACE\cite{brachmann2023accelerated}, except that the network width is adjusted to match the input dimension of concatenated encoding. In addition, we use more residual blocks and increase the hidden size of the residual block for large outdoor scenes such as Cambridge and Aachen to increase model capacity, while still maintaining a comparable map size as baseline methods. We also tried concatenating the Superpoint~\cite{detone18superpoint} descriptor to the original ACE local encoder for the Aachen dataset to provide a more discriminative local descriptor. 

\customparagraph{Training}
Most of the training parameter choices are the same as ACE, but we use larger buffer sizes for larger scenes, because there is more training data to be cached. In addition, we also use a larger batch size. As shown in Sec.~\ref{sec:SCR}, the reprojection supervision acts as an implicit triangulation. Therefore, it is desirable to have multiple observations of the same point in one batch to get stable and accurate supervision. In order to cache these larger buffers, we use distributed training with multiple GPUs. 
Specifically, we use a batch size of 160K and a training buffer size of 64M for the Cambridge dataset, a batch size of 320K and a training buffer size of 128M for Aachen and i19. For the Superpoint~\cite{detone18superpoint} version on Aachen, we also perform importance sampling according to its corner detection likelihood in order to select more salient structures. 
We train 30k iterations for Cambridge and 100k iterations for Aachen and i19.

\subsection{Evaluation Results}
\customparagraph{7 Scenes and 12 Scenes}
As indicated in Tab.~\ref{tab:results_single}, our approach retains the benefits of accuracy and compact map size observed in SCR methods when applied to small room-scale scenes.

\customparagraph{Integrated Rooms}
As shown in Tab.~\ref{tab:results_id} and Tab.~\ref{tab:results_i}, previous SCR methods need a much larger map size, or demand an ensemble of networks in order to achieve satisfactory performance on large indoor scenes. Our method achieves comparable performance by a single model with a much smaller total map size.  During test time, we only need to query a single model instead of all the ensemble models, which also makes our method more efficient and practical.

\customparagraph{Cambridge Landmarks}
This real-world outdoor dataset can fully demonstrate the advantages of our method. As shown in Tab.~\ref{tab:results_cam}, our method significantly outperforms state-of-the-art SCR methods~\cite{brachmann2023accelerated,brachmann2017dsac} and closes the gap with FM methods~\cite{sarlin21pixloc,sarlin2019coarse}. Particularly, the largest scene in this dataset, GreatCourt, is very challenging for SCR methods, but our method can still achieve comparable performance to FM methods with a small model size.

\customparagraph{Aachen Day}
\begin{table*}[]
  \centering
  \small
  \begin{tabular}{@{}lccccc@{}}
  \toprule
  Method & w / Depth & Size & 0.25m,2$^\circ$ & 0.5m,5$^\circ$ & 5m,10$^\circ$ \\ \midrule
  ESAC $\times$ 50 \cite{Brachmann2019ESAC} & Yes & 1400MB& 42.6\%  & 59.6\%  & 75.5\%  \\ 
  \midrule
  ACE~\cite{brachmann2023accelerated} $\times$ 4 & No &16MB & 0.0\% & 0.5\% & 3.8\% \\
  ACE~\cite{brachmann2023accelerated} $\times$ 50 & No & 205MB & 6.9\% & 17.2\% & 50.0\%                \\ 
  GLACE (Ours) & No & 27MB &   8.6\% & 20.8\%  &  64.0\% \\ 
  GLACE (Ours, SuperPoint~\cite{detone18superpoint})& No & 23MB& 9.8\% & 23.9\% &  65.9\%\\
   \bottomrule
  \end{tabular}
  \caption{\textbf{Aachen Day evaluation.} We compare the accuracy on Aachen Day dataset~\cite{sattler2018benchmarking,sattler2012image}. }
  \label{tab:result_aachen}
  \end{table*}
We also evaluate our method on the Aachen dataset. The challenges of this dataset are not only the scale but also the sparsity. There are only about 4K discrete images for a city-scale scene, while the other datasets consist of several sequences with thousands of images for a small scene. Previous methods~\cite{Brachmann2019ESAC} usually rely on the ground truth scene coordinate supervision, however, we can still achieve comparable results without ground truth scene coordinate supervision and a much smaller map size. Other methods~\cite{brachmann2023accelerated} that also feature small map sizes and no scene coordinate supervision will fail with a similar map size as ours. They cannot achieve a similar performance even with an ensemble of 50 models. In addition, we also tried concatenating SuperPoint~\cite{detone18superpoint} features to the original ACE~\cite{brachmann2023accelerated} local features to increase the discriminative power and achieved better performance with smaller map size.

\subsection{Ablation Study}

\begin{table*}[h]
  \centering
  \small
  \begin{tabular}{lccccccc}
  \toprule
  \multirow{2}{*}{Scene} & \multirow{2}{*}{ACE~\cite{brachmann2023accelerated}}& \multirow{2}{*}{Poker (ACE $\times$ 4)} & \multicolumn{5}{c}{GLACE (Ours)} \\
  \cmidrule{4-8}
  && &Identity& K=4 & K=32& K=128 & Diffusion\\
  \midrule
      GreatCourt  & 43/0.2&   28/0.1  & 32/0.2 &27/0.1&23/0.1&23/0.2 & 19/0.1 \\
  KingsCollege    & 28/0.4&  18/0.3  & 30/0.4 &18/0.3&19/0.3&22/0.4 & 19/0.3 \\
  OldHospital     & 31/0.6& 25/0.5   & 34/0.6 & 21/0.4&20/0.4&21/0.4 & 17/0.4 \\
  ShopFacade     & 5/0.3&   5/0.3  & 13/0.5 &5/0.2&6/0.2&6/0.3& 4/0.2 \\
  StMarysChurch   & 18/0.6& 9/0.3   & 103/2.1 &9/0.3&10/0.3&10/0.4 & 9/0.3 \\
  \midrule
  Average         & 25/0.4&  17/0.3   & 42/0.8 & 16/0.3 &15/0.3&17/0.3 & 14/0.3 \\
  \bottomrule
  \end{tabular}
  \caption{\textbf{Ablation of Global Encoding.} Performance of GLACE on the Cambridge Landmarks~\cite{kendall2015posenet} with different kinds of global encoding input. We report median rotation and position errors.}
  \label{tab:gfeat-ablation}
   \vspace{-1mm}
\end{table*}

\customparagraph{Feature Diffusion}
Tab.~\ref{tab:gfeat-ablation} compares different kinds of global encoding input on Cambridge Landmarks~\cite{kendall2015posenet} and shows the effectiveness of our feature diffusion technique. When we directly concatenate the global encoding to the local encoding, the performance suffers from overfitting, especially apparent for simple scenes like StMarysChurch. 
If we use K-Means to cluster the global encoding to certain discrete center values, we can explicitly force the grouping of the reprojection constraints. However, it is non-trivial to choose a suitable number of clusters, which may require a lot of tuning. In contrast, our feature diffusion technique achieves the best performance and additionally avoids tuning any hyperparameter.

\customparagraph{Decoder}
\begin{figure}[h]
  \centering
   \includegraphics[width=1.0\linewidth]{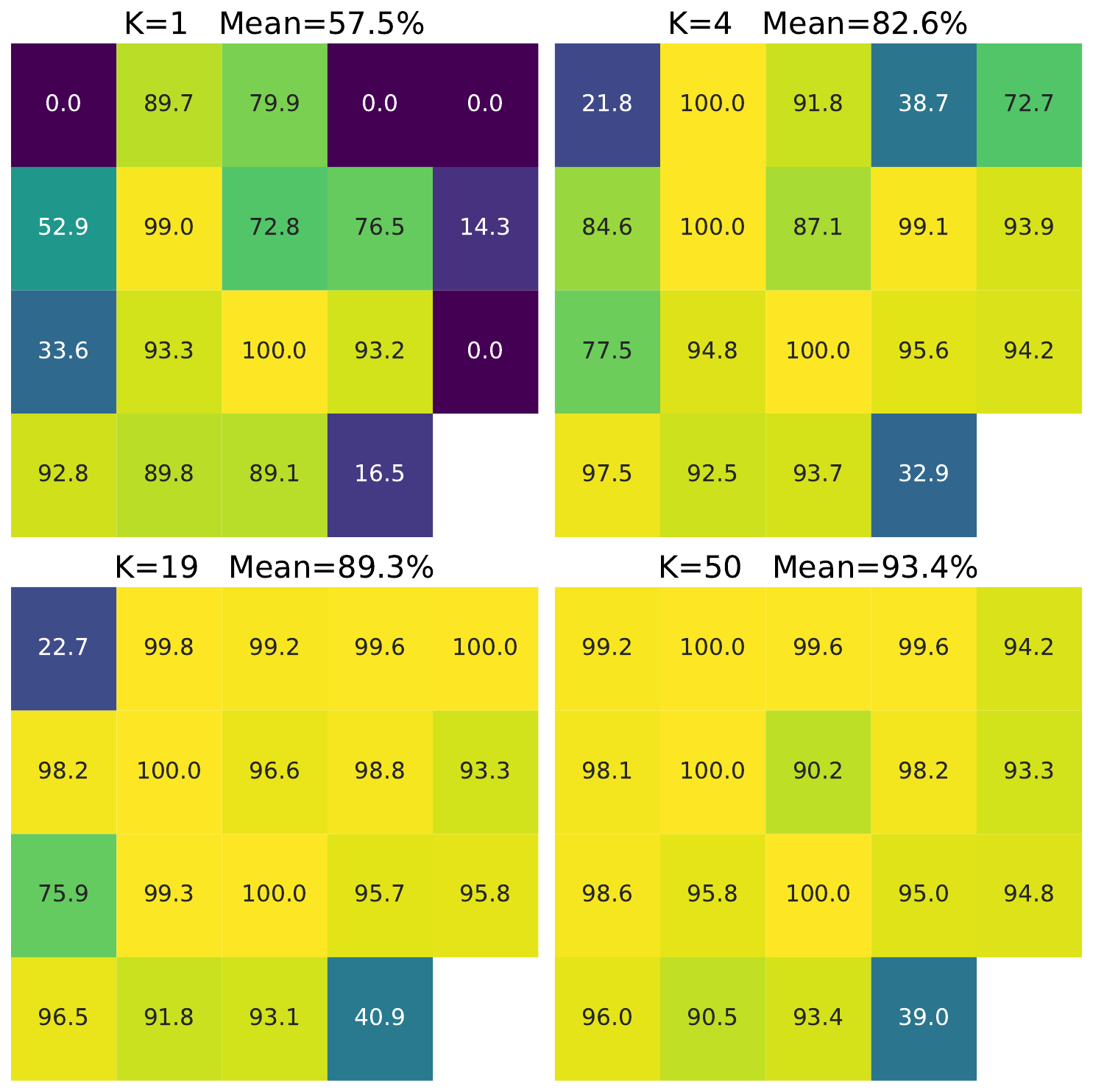}
   \caption{\textbf{Ablation of Decoder.} We compare the percentage of frames below a 5cm, 5$^\circ$ pose error for each room in the i19 integrated dataset.}
   \label{fig:decoder_i19}
   \vspace{-1mm}
\end{figure}
In Fig~\ref{fig:decoder_i19}, we show the performance of our method with different numbers of decoder clusters $K$ on the i19 dataset with SfM ground truth. When $K=1$, which is equivalent to the original ACE~\cite{brachmann2023accelerated} decoder, the network has an unimodal prior, which only learns the center scenes well and almost completely fails on several border scenes that are away from the center. When we increase the number of decoder clusters, the model is allowed to better parameterize a multimodal distribution and have increasing performance in border scenes. 
Note that different from the ensemble methods~\cite{brachmann2023accelerated} that splits the scene into clusters and trains multiple models,
our method of increasing the number of decoder clusters only needs to add a few output channels for the last linear layer and shows no significant increase in inference time and model size. 

%% file: sec/5_conclusion.tex
\section{Conclusion}\label{sec:conclusion}

In this paper, we have presented GLACE, a novel scene coordinate regression method that is able to work on large-scale scenes with a single network and without ground truth scene coordinate supervision. 
We propose a feature diffusion technique that effectively utilizes co-visibility information in the form of global encoding from image retrieval networks, to implicitly group the reprojection constraints and avoid overfitting to trivial solutions. We also propose a position decoder to effectively parameterize output coordinates in large-scale scenes. 
We believe that our insights and technical solutions are also applicable to other SCR methods to improve their performance on large-scale scenes. 

%% file: sec/X_suppl.tex
\clearpage
\setcounter{section}{0}
\renewcommand{\thesection}{\Alph{section}}
\maketitlesupplementary

\begin{figure*}[t]
\centering
    \includegraphics[width=\linewidth]{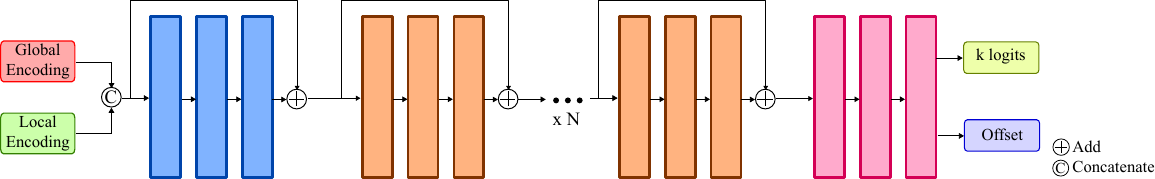}
\caption{Detailed structure of our fully-connected network architecture for GLACE. In the beginning, a residual block (blue) transforms the concatenation of global and local encodings, which is followed by $N$ sequential residual blocks (orange). Finally, three fully-connected layers (pink) are applied to get the $k$ logits and offset for estimating the 3D position. }
\label{fig:network}
\end{figure*}

\section{Network Architecture}
The detailed structure of our method is shown in Fig.~\ref{fig:network}. Our network has a fully-connected network architecture. By default, we set the hidden size $h$ of all layers as 768, which is the size of the concatenation of local encoding (512) and global encoding (256). 

In the beginning, the network takes as input the concatenation of local encoding and global encoding, and transforms it with a residual block which has 3 fully-connected layers. 
Then $N$ residual blocks with 3 fully-connected layers are sequentially applied, where we choose $N$ based on the scene scale. 
Specifically, we set $N=1$ for all indoor scenes, and $N=2$ for Cambridge Landmarks~\cite{kendall2015posenet}.
For Aachen Day~\cite{sattler2018benchmarking,sattler2012image} dataset, we set $N=3$ and double the hidden size $h$,~\ie 1536, for the second layer of each residual block. 
In addition, for evaluation of Aachen Day~\cite{sattler2018benchmarking,sattler2012image} dataset with additional SuperPoint~\cite{detone18superpoint} feature, we set $N=2$ and do not double the hidden size $h$ in order to maintain a similar map size. 
Finally, we apply 3 fully-connected layers to get $k$ logits $\{s_i\}$, one for each cluster center, and one homogeneous coordinate with parameters $\Dot{d},\hat{w}$ to define an offset.  The final 3D coordinate $\hat{y}$ is estimated with Eq.~8 in the main paper. 

To get the $k$ cluster centers from training data, we cluster training camera positions with K-Means++~\cite{kmeans++}. We set $k = 50$ for scenes that have a more multimodal distribution, including integrated rooms and Aachen Day dataset~\cite{sattler2018benchmarking,sattler2012image}, and $k = 1$ for scenes that have a more unimodal distribution, including individual scenes in the 7 Scenes~\cite{shotton2013scene}, 12 Scenes~\cite{valentin2016learning} and Cambridge Landmarks~\cite{kendall2015posenet}. 

\section{Experiment Details}
Following~\cite{brachmann2023accelerated}, we allocate a training buffer on the GPU, which stores local encodings and corresponding metadata,~\ie image indices and ground truth poses. This buffer is filled by iterating over the training images. Each image is first converted to grayscale and then subjected to a series of data augmentations: random scaling between $\frac{2}{3}$ and $\frac{3}{2}$, brightness and contrast jitter by $10\%$, and random rotations up to a maximum of $15^\circ$. From each augmented image, we extract and uniformly sample 1024 local encodings. For the version using SuperPoint~\cite{detone18superpoint}, importance sampling based on corner detection probability is employed instead. We also continuously update the training buffer during each training iteration when the number of training images is large.

Global features for each training image are extracted without any data augmentation and stored in a lookup table to avoid unnecessary duplication. During each training iteration, a batch of local encodings is randomly selected from the training buffer. Corresponding global encodings are then retrieved based on the image index. For these global encodings, we add Gaussian noise with a standard deviation of $\sigma = m = 0.1$, where $m$ is the margin used in the triplet margin loss by the global feature extractor~\cite{r2former}. Subsequently, the global encodings are normalized back to the unit sphere. 

We use AdamW~\cite{Loshchilov2019DecoupledWD} optimizer with a One Cycle learning rate scheduler~\cite{Smith2019cycliclr} that increases the learning rate from $2\cdot 10^{-4}$ to $5\cdot 10^{-3}$ and then decreases to $2\cdot 10^{-8}$. The detailed mapping times and number of GPUs for training is shown in Tab.~\ref{tab:mapping_time}. %

During evaluation, we use a 10px inlier threshold and 64 RANSAC hypotheses for all experiments, except that we use 3200 RANSAC hypotheses for Aachen Day~\cite{sattler2018benchmarking,sattler2012image} dataset to match the number of RANSAC hypotheses of the ACE~\cite{brachmann2023accelerated} $\times$ 50 baseline. 

\begin{table}[]
\centering
\footnotesize
\begin{tabular}{lcc}
\toprule
 Scene & Number of GPUs                        & Mapping Time  \\ \midrule
 7 / 12 Scenes~\cite{shotton2013scene,valentin2016learning}& 1 & 6 min\\
 Cambridge~\cite{kendall2015posenet} & 4 & 20 min\\
 i12 / i19& 8 & 1 h 50 min\\
 Aachen Day~\cite{sattler2018benchmarking,sattler2012image}&  8  & 2 h 30 min\\
\bottomrule
\end{tabular}
\caption{Mapping Times of our method on different scenes. We use Nvidia Quadro RTX 6000 GPUs in experiments.}
\label{tab:mapping_time}
\end{table}

\begin{figure}[t!]
\centering
    \includegraphics[width=\linewidth]{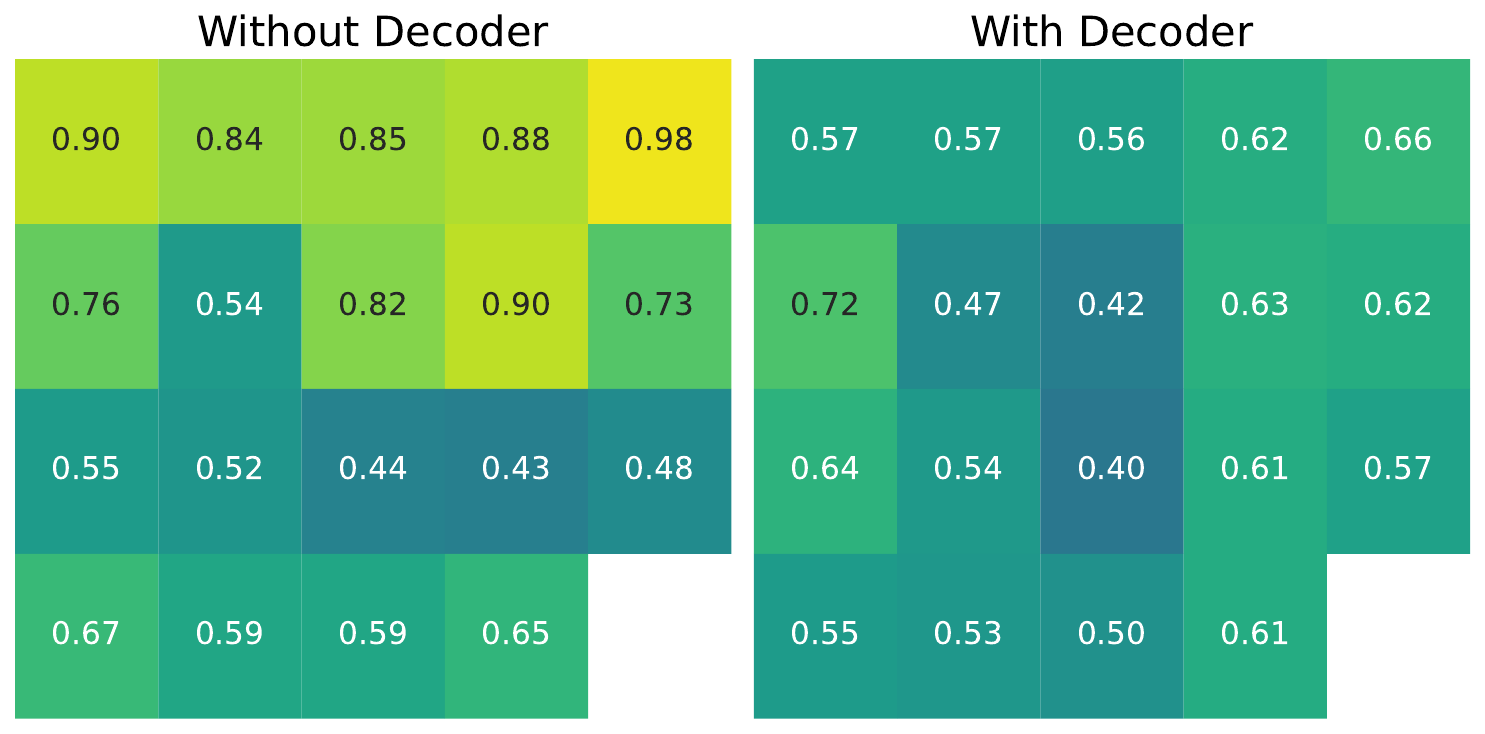}
\caption{Comparison about mean absolute error of pixel location prediction in our 2D toy example.}
\label{fig:toy}
\end{figure}

\section{Position Decoding in 2D Toy Example}
We designed a simplified 2D toy example to show the effect of our position decoder. We randomly select 19 images from the 7 Scenes~\cite{shotton2013scene} and 12 Scenes~\cite{valentin2016learning} datasets and place them in a grid with a similar layout as the i19 scene. 
The images are resized and cropped to a size of 480 x 640 for convenient batch processing. We use the same pretrained ACE~\cite{brachmann2023accelerated} encoder and train the MLP head with similar architecture, except that the output coordinate is now 2D instead of 3D. We use 19 decoder cluster centers, which are actually the centers of the 19 images. The output coordinate is directly supervised by the ground truth pixel location.
Fig.~\ref{fig:toy} shows that, even for this simple example with strong supervision, our position decoder can allow the model to fit the training data with a multi-modal output distribution better.

\section{Reconstruction Visualization}
\begin{figure*}[h]
\centering
\begin{subfigure}{0.5\textwidth}
    \centering
    \includegraphics[width=0.9\textwidth]{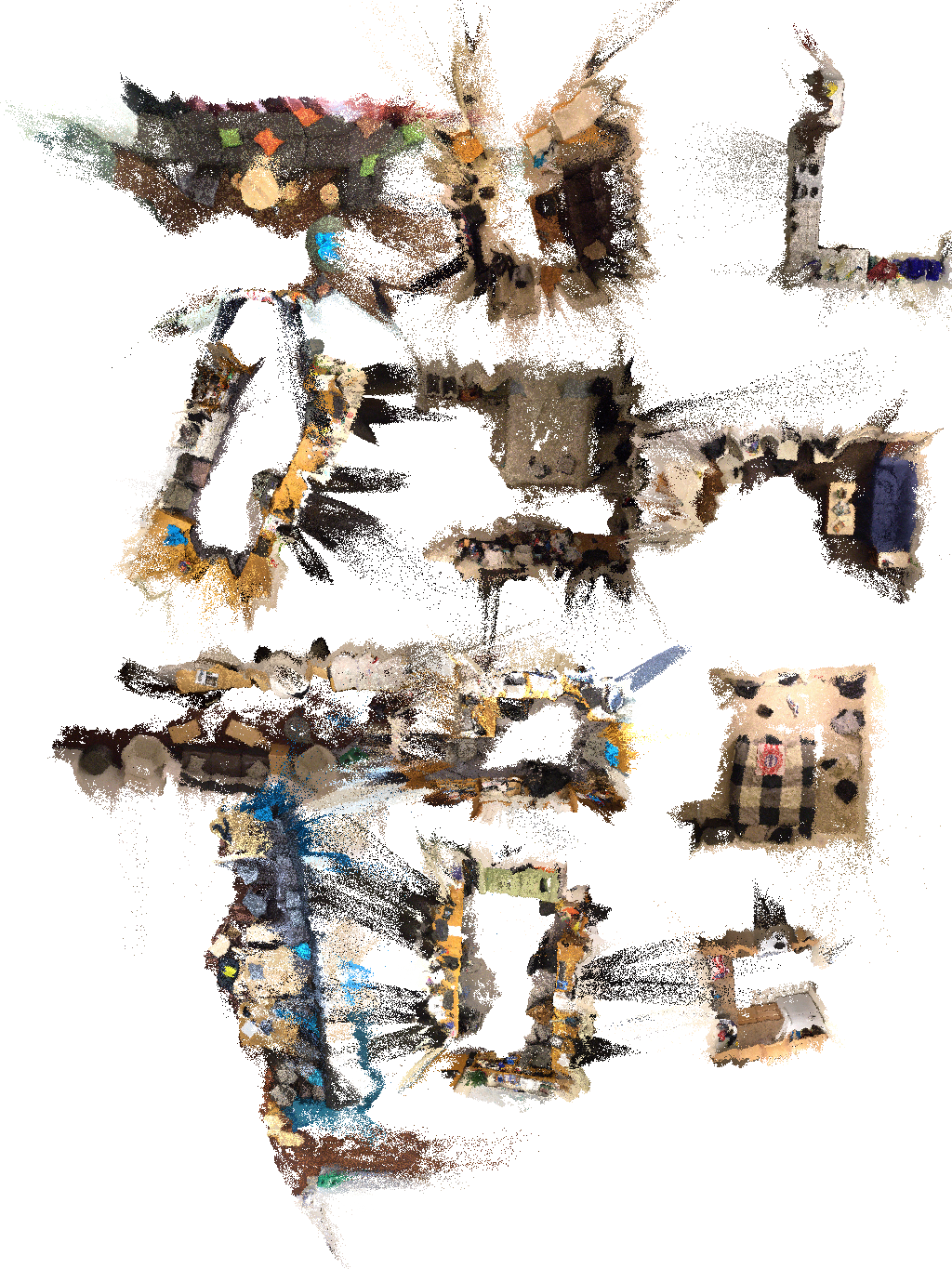}
    \caption{i12}
\end{subfigure}%
\begin{subfigure}{0.5\textwidth}
    \centering
    \includegraphics[width=0.9\textwidth]{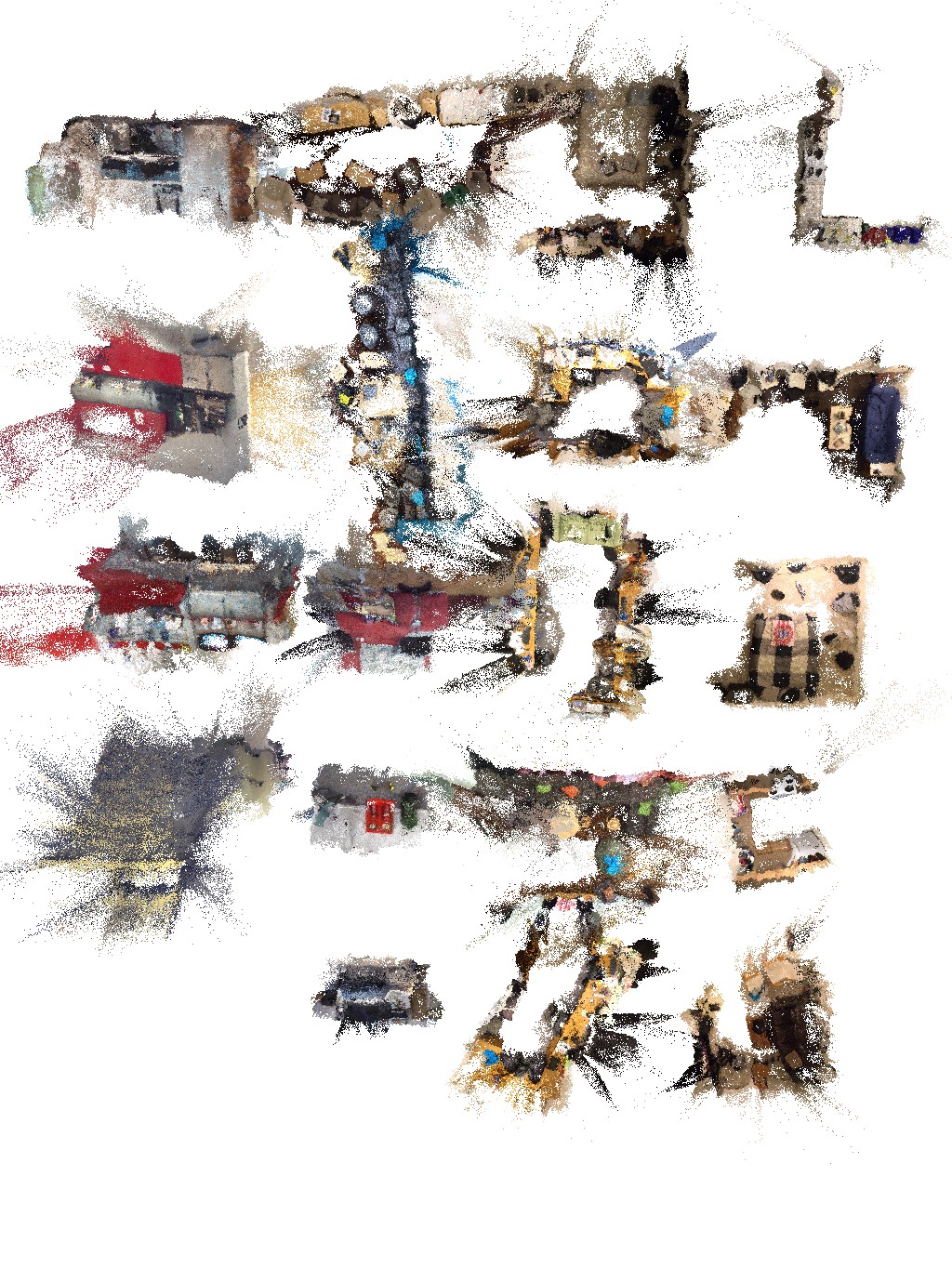}
    \caption{i19}
\end{subfigure}%
\caption{Reconstructions of integrated rooms.}
\label{fig:visualize_i}
\end{figure*}

\begin{figure*}[b]
\centering
    \includegraphics[width=\textwidth]{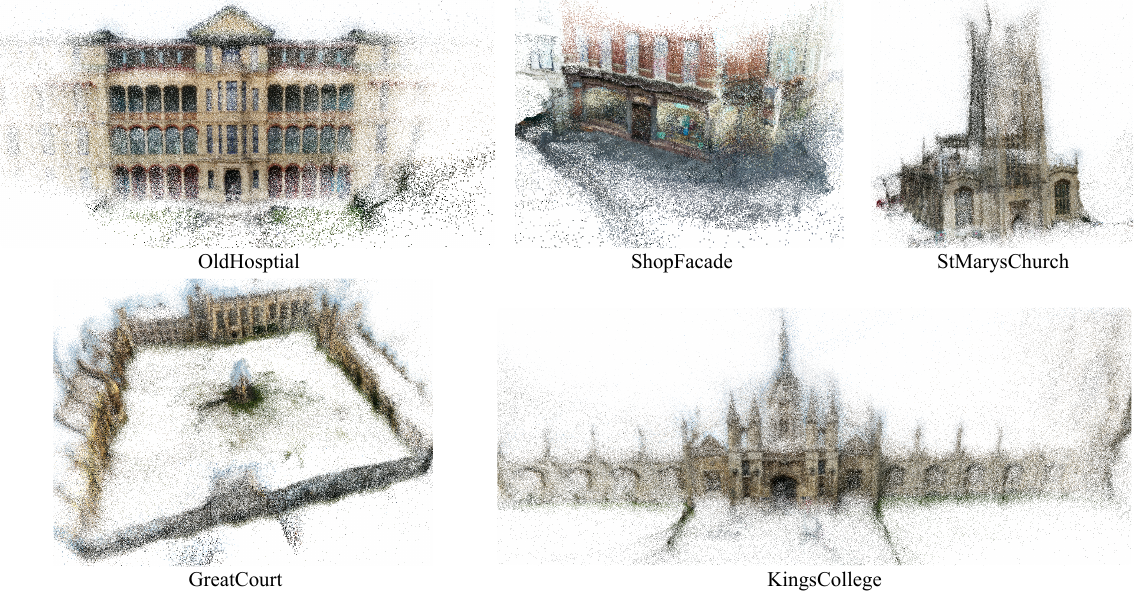}
\caption{Reconstructions of Cambridge Landmarks~\cite{kendall2015posenet}. }
\label{fig:visualize_cambridge}
\end{figure*}

\begin{figure*}[b]
\centering
    \includegraphics[width=\textwidth]{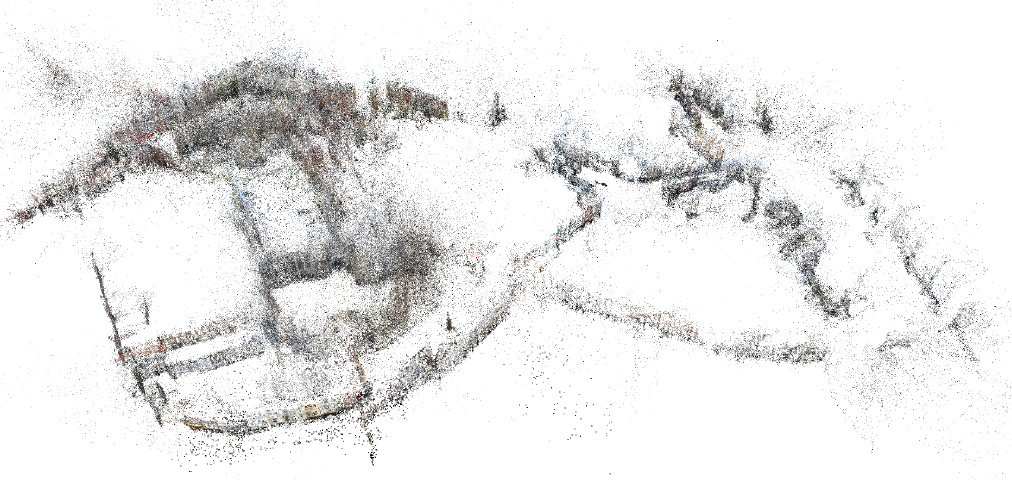}
\caption{Reconstrucion of Aachen Day dataset~\cite{sattler2018benchmarking,sattler2012image}. }
\label{fig:visualize_aachen}
\end{figure*}

In Fig.~\ref{fig:visualize_i}, \ref{fig:visualize_cambridge} and \ref{fig:visualize_aachen}, we visualize the implicit reconstructions by accumulating the predicted 3D scene coordinates of the training images, and filter the outliers according to a 5px reprojection error threshold. The point cloud color is obtained from the center pixel of each image patch. As we can see, the implicit triangulation allows the model to learn meaningful 3D structures from reprojection loss only.

\clearpage

